\documentclass[10pt,twocolumn,letterpaper]{article}

\usepackage[pagenumbers]{cvpr} 

\usepackage[normalem]{ulem}
\usepackage{bm}
\usepackage{multirow}  
\usepackage{makecell}  
\usepackage{booktabs}   
\usepackage{amssymb} 
\usepackage{pifont}
\definecolor{skyblue}{HTML}{DEEBFE}
\usepackage{nameref}

\usepackage{tikz}
\newcommand*\Circled[1]{
	\tikz[baseline=(char.base)]{\node[
        shape=circle, draw=none,  thick, 
        fill=gray!40,inner sep=0.9pt] (char) 
    {\textcolor{black}{#1}}; 
}}
\newcommand{\xmark}{\ding{55}}  
\newcommand{\cmark}{\ding{51}}

\newcommand{\abbr}{AdaRing}

\definecolor{cvprblue}{rgb}{0.21,0.49,0.74}
\usepackage[pagebackref,breaklinks,colorlinks,allcolors=cvprblue]{hyperref}

\def\paperID{*****} 
\def\confName{CVPR}
\def\confYear{2026}

\title{AdaRing: Towards Ultra-Light Vision-Language Adaptation via Cross-Layer Tensor Ring Decomposition}

\author{ Ying Huang$^{\text{1}}$, Yuanbin Man$^{\text{1}}$, Wenqi Jia$^{\text{1}}$, Zhengzhong Tu$^{\text{2}}$, Junzhou Huang$^{\text{1}}$, Miao Yin$^{\text{1}{\dagger}}$\\
{$^\text{1}$University of Texas at Arlington, $^\text{2}$Texas A\&M University}\\
{\tt\small \texttt{\{ying.huang, yuanbin.man, wenqi.jia\}}@uta.edu, tzz@tamu.edu, } \\
{\tt\small  \texttt{\{jzhuang, miao.yin\}}@uta.edu}
}

\begin{document}
\maketitle
\begin{abstract}

\let\thefootnote\relax\footnotetext{\hspace{-5mm}$^\dagger$Corresponding author.}Adapter-based fine-tuning has gained remarkable attention in adapting large pre-trained vision language models (VLMs) for a wide range of downstream tasks efficiently. In this paradigm, only the inserted adapters are fine-tuned, without the need for training the original VLM backbone. 
Existing works scale adapters by integrating them into every layer of VLMs to increase the capacity of adapters. 
However, these methods face two primary limitations: 1) limited compression rate due to ignoring cross-layer redundancy, and 2) limited representational capacity across homogeneous adapters. 
In this paper, we propose a novel vision-language fine-tuning framework based on cross-layer tensor ring decomposition (TRD) with the integration and collaboration of diverse adapters, called \abbr, achieving ultra-light parameter-efficient adaptation of VLMs on various tasks. To remove the high redundancy that exists among adapters across layers, we exploit the tensor-level low-rankness to formulate adapters as layer-shared tensor cores and layer-specific slices. Moreover, guided by generalization-aware fine-tuning, diverse rank-driven adapters cooperate to handle tasks that require different representations.
Our experiments show that the proposed \abbr~achieves the state-of-the-art performance while reducing average training parameters by 90\%.
\end{abstract} 
\section{Introduction}
Vision-Language Models (VLMs) adaptation~\cite{zhou2022learning, zhou2022conditional, cho2024cat, jin2025lor} has gained substantial attention in the fields of computer vision (CV) and natural language processing (NLP), as a means of adapting VLMs, e.g., CLIP~\cite{jia2021scaling} and ALIGN~\cite{radford2021learning}, pre-trained from web-scale image-text pairs into multiple downstream tasks.
Despite the effectiveness of representations that benefit from large-scale pre-training, fine-tuning VLMs for downstream tasks still faces several challenges, especially the challenge of \textit{massive fine-tuning parameters}, which results in heavy computational costs and a significant memory burden~\cite{peng2024parameter, wu2025longvituinstructiontuninglongform, jiao2024visual}.

\textit{Adapter-based fine-tuning}~\cite{gao2024clip, zhao2024dynamic} is the mainstream paradigm to mitigate this issue, where small trainable adapters, typically implemented as two linear layers, are inserted into the final layer of the VLM. During training, only these lightweight modules are fine-tuned, while the rest of the parameters in the VLM backbone remain frozen. This approach drastically reduces the number of trainable parameters. However, confining the adaptation to only the final layer severely limits the model's capacity to capture the complex information required by real-world vision-language tasks.

\begin{figure}[t]
    \includegraphics[width=0.48\textwidth]{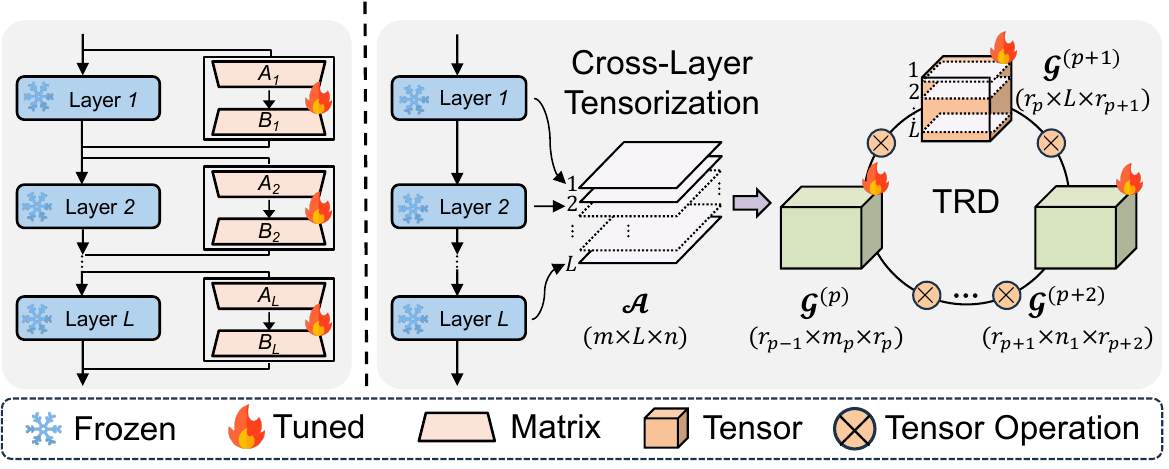}
    \vspace{-5mm}
    \caption{(Left) Existing methods utilize regular matrix decomposition layer by layer, leading to significant redundancy and limited representation capability. (Right) Our \textbf{\abbr}~achieves ultra-light adaptation via cross-layer tensor ring decomposition (TRD). }
    \label{fig:existing_methods_new}
\end{figure}

\begin{figure}[t]
    \includegraphics[width=0.48\textwidth]{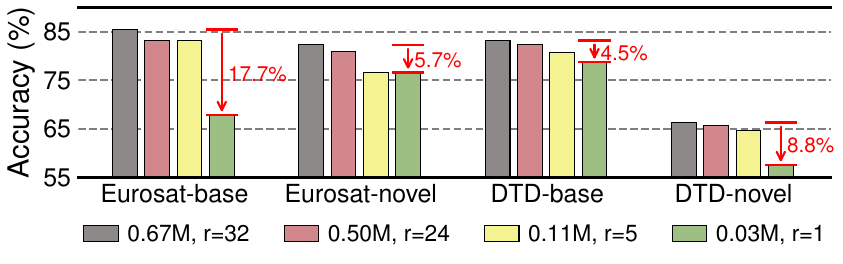}
    \vspace{-5mm}
    \caption{Matrix decomposition-based approaches (e.g., MMA) suffer from rapid performance degradation as the number of trainable parameters decreases on the DTD and EuroSAT datasets.}
    \label{fig:varied_rank}
\end{figure}

To address this limitation, existing methods scale adapters beyond the final layer, either \textit{vertically}, by inserting them into every transformer block~\cite{chen2022adaptformer, yang2024mma}, or \textit{horizontally}, by introducing multiple adapters within a single block~\cite{yu2024boosting}. While these schemes improve performance, they still face two main limitations. 
\uline{1) Inadequate compression rate.}
Based on the hypothesis that the adapter weight has a low ``intrinsic rank'', current methods leverage \textit{low-rank matrix decomposition}~\cite{hu2022lora} for every adapter independently to reduce training parameters in Figure \ref{fig:existing_methods_new}. 
However, such matrix-level decomposition methods still have a significant accuracy decrease with insufficient compression rates as shown in Figure \ref{fig:varied_rank}. 
Moreover, these approaches ignore the redundancy across adapters of different layers, resulting in a large number of training parameters, especially in the large VLM with deep layers.
\uline{2) Limited representational capacity across homogeneous adapters.}
Most existing methods introduce multiple uniform, homogeneous adapters within a single layer, which tend to learn similar features during fine-tuning. 
This lack of diversity makes it challenging to meet the varied demands of different features for tasks, hindering
the model’s ability to generalize across varied tasks~\cite{wang2024hmoe}. 
\textit{Quantitative} analysis is presented in the following Section \ref{sec:motivation}.

\begin{figure*}[t]
    \centering
    \includegraphics[width=\textwidth]{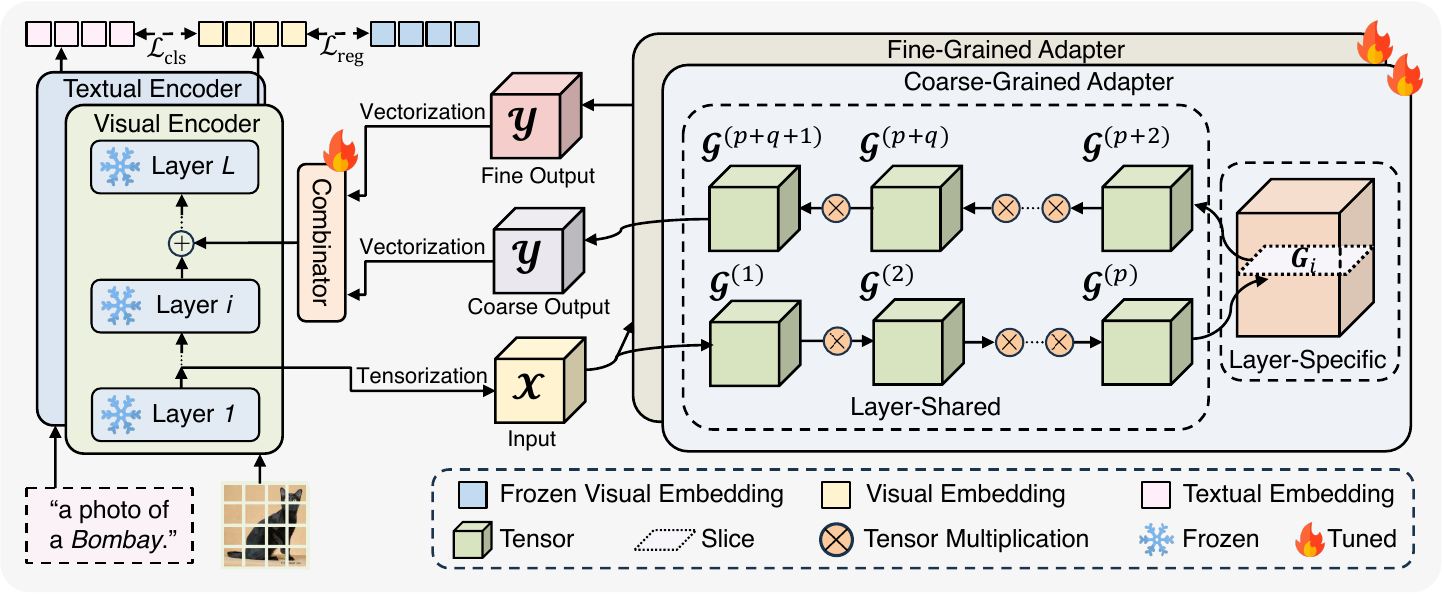}
    \vspace{-5mm}
    \caption{The framework of \abbr. With image and text as inputs,  \abbr~equips the $i$-th encoder layer with fine-grained adapter and coarse-grained adapter to handle tasks that require distinct representational capacities. The adapter consists of the layer-shared tensor cores and the layer-specific slice to extract redundancy across adapters and simultaneously preserve layer-specific information. During fine-tuning, the textual and visual encoders remain entirely frozen, while the adapter is trained to minimize cross-entropy loss $\mathcal{L}_{\mathrm{cls}}$ combined with a generalization-aware regularization term $\mathcal{L}_{\mathrm{reg}}$, retaining strong discrimination while enhancing generalization in downstream tasks.}
    \label{fig:adaRing}
\end{figure*}

To address those limitations, we propose, \abbr, an ultra-light vision-language adaptation framework using
cross-layer tensor ring decomposition. Specifically, instead of applying low-rank approximation to each layer’s weight matrix independently, we consider stacking the adapters across layers into a higher-order tensor, and then compressing it to reduce both intra-layer and cross-layer redundancy. Fortunately, \textit{tensor decomposition}~\cite{yin2022hodec, xiao2023haloc} offers a principled way to achieve this: it decomposes the original large, high-dimensional weight tensor into a sequence of compact tensor cores, leading to a significant reduction in training parameters. Moreover, the rank of each tensor core provides a mechanism to modulate the model's representational specializations~\cite{xiao2023comcat,wang2024hmoe}.
Motivated by these advantages, we conduct an in-depth exploration of tensor decomposition-based VLM adaptation with diverse adapters. 
First, we investigate the redundancy of cross-layer adapters, which opens opportunities for achieving an ultra-light adapter based on cross-layer tensor decomposition.
Second, we explore different variants of tensor decomposition, showing that the adapter of tensor ring~\cite{zhao2016tensor} format has more powerful and separate representation abilities than other format, as observed in our experiments with CLIP. 
Third, to design adapters that can handle various downstream tasks effectively, we equip CLIP with diverse adapters and achieve better performance than homogeneous adapters.

Building on these insights, we propose \abbr, a novel fine-tuning framework based on cross-layer tensor ring (TR) decomposition with the integration and collaboration of diverse adapters, to achieve extremely parameter-efficient adaptation of VLMs for varied tasks. Our key contribution can be summarized as follows:
\begin{itemize}
\item We creatively reduce the cross-layer redundancy by exploiting the tensor-level low-rankness, stacking all adapters across different layers, and decomposing it into a high-dimensional tensor with layer-shared and layer-specific cores. Our framework achieves ultra-light adaptation on real-world tasks. 

\item We equip VLMs with diverse adapters guided by rank to handle varied tasks, and a learnable combinator is then designed to coordinate their collaboration adaptively. We propose generalization-aware fine-tuning to improve performance on previously unseen data.

\item We conduct experiments on extensive downstream fine-tuning tasks. Our \abbr~achieves the best average performance over 11 datasets and outperforms the previous state-of-the-art MMA~\cite{yang2024mma} under almost all few-shot scenarios in the OxfordFlowers dataset. Moreover, \abbr~reduces the average training parameters by 90\%, highlighting superior performance and efficiency in practice.
\end{itemize}

\section{Related Work}

\textbf{Parameter-efficient Fine-tuning.} Parameter-efficient fine-tuning technique aims to transfer pretrained models, e.g., CLIP~\cite{jia2021scaling}, to downstream tasks by only fine-tuning a few trainable parameters for efficiency~\cite{chen2025sensitivity, pujol2025sparse, luo2025tr}. In general, it can be realized via either prompt tuning, adapter tuning, or reparameterization tuning. 
\textbf{Prompt tuning} involves incorporating handcrafted or learnable prompts to facilitate the fine-tuning of a pre-trained model~\cite{zhou2022learning, zhou2022conditional, yao2023visual}. PromptKD leverages a two-stage prompt tuning process to encourage the knowledge distillation from a large teacher model to a lightweight target model~\cite{li2024promptkd}. 
Instead of fine-tuning input prompts, \textbf{adapter tuning} proposes to insert some small modules into the original model, and only these inserted modules are optimized during fine-tuning~\cite{gao2024clip}. Existing adapter tuning methods are mainly built upon \textit{layer-wise} adapters~\cite{zhao2024dynamic}. For exmaple, MMA~\cite{yang2024mma} adds the new adapter into higher layers of models to provide alignment between vision-language representations. In addition, \cite{yu2024boosting} introduces MoE structure onto each layer of frozen CLIP, enabling the dynamic expansion of a pre-trained CLIP model. 
However, these methods ignore the parameter redundancy among adapters of different layers, leading to a substantial increase in computational burden as model layers increase.

\textbf{Low-rank Matrix \& Tensor Decomposition of Weight Matrices.}
Low-rank decomposition is a powerful compression tool that can reduce model size and bring considerable speedup. It can broadly be categorized into matrix decomposition and tensor decomposition. 
\textbf{Matrix decomposition} factorizes the large weight matrix into small matrix components~\cite{jaderberg2014speeding, hayou2024lora+, he2022towards}. However, these approaches do not fully exploit the inherent spatial low-rankness of the weight, causing significant accuracy loss with limited compression ratios.  To address these limitations, \textbf{tensor decomposition}, such as Tucker~\cite{tucker1963implications} and Tensor Train ~\cite{harshman1970foundations, yang2024adazeta, yang2024loretta}, factorizes the original high-order tensor in the high-dimensional space~\cite{wang2022exploring}, which can bring impressive compression performance.
In this paper, we efficiently tensorize and decompose the weight matrices of the adapters during fine-tuning VLMs. Moreover, instead of applying tensor decomposition to each layer-wise adapter independently, we utilize \textit{cross-layer} tensor decomposition to simultaneously reduce inter-layer and intra-layer redundancy.

\section{Background}

\subsection{Notation}
Throughout this paper, vectors, matrices, and tensors are denoted by bold lowercase letters, bold uppercase letters, and bold calligraphic letters, respectively, e.g., $\bm{a}$, $\bm{A}$ and $\bm{\mathcal{A}}$. Also, $\bm{\mathcal {A}}_{(i_1, \cdots, i_d)}$ denotes the single entry of $d$-order tensor $\bm{\mathcal{A}}$. 

\subsection{Adapter Tuning Basics} 

Adapter tuning achieves comparable performance to fine-tuning VLM (CLIP~\cite{jia2021scaling, esfandiarpoor-etal-2024-clip} in this work) for many downstream tasks by keeping the backbone visual encoder $\mathcal{V}$ and 
textual encoder $\mathcal{T}$ frozen and only fine-tuning additional inserted visual adapter and textual adapter. 

Recent works insert adapters into each layer to increase the capacities, thereby improving fine-tuning performance. Moreover, in order to alleviate the training costs caused by substantial fine-tuning parameters, they use two sequential low-rank matrices $\bm{A}_{l}\in \mathbb{R}^{I \times r} $ and $\bm{B}_{l}\in \mathbb{R}^{r \times O}$ as each layer-wise adapter $\mathcal{F}_{l}$~\cite{hu2022lora}. 
The forward computation for the $l$-th layer is written as:
\begin{equation}
\begin{aligned}
\boldsymbol{y}^{v} &= \mathcal{V}_l(\boldsymbol{x}^{v}) + \alpha \, \bm{A}_l^{v}\bm{B}_l^{v}\boldsymbol{x}^{v}, \\
\boldsymbol{y}^{t} &= \mathcal{T}_l(\boldsymbol{x}^{t}) + \beta \, \bm{A}_l^{t}\bm{B}_l^{t}\boldsymbol{x}^{t},
\end{aligned}
\label{eq:adapter}
\end{equation}
where $\mathcal{V}_l$, $\boldsymbol{x}^{v}$ and $\boldsymbol{y}^{v}$ denote the $l$-th visual encoder layer, corresponding input and output visual embeddings, respectively; $\mathcal{T}_l$, $\boldsymbol{x}^{t}$ and $\boldsymbol{y}^{t}$ represent the $l$-th textual encoder layer, corresponding input and output textual embeddings, respectively; $\alpha$ and $\beta$ are coefficients to merge embeddings produced by the frozen encoder layer and adapter. Finally, the specific image feature $f^{v}$ and text feature $f^{t}$ are obtained from the last layer. The adapters are trained to maximize the similarity of $f^{v}$ and the corresponding $f^{t}$.

\subsection{Tensor Ring Decomposition Basics}

Given a $d$-order tensor $\bm{\mathcal{A}} \in \mathbb{R}^{n_1 \times n_2 \times \cdots \times n_d}$, tensor ring decomposition (TRD)~\cite{zhao2016tensor} can represent it with a sequence of 3-order core tensors $\bm{\mathcal{G}}^{j} \in \mathbb{R}^{R_{j-1} \times n_j \times R_j}$ as follows:
\vspace{-1mm}
\begin{equation}
\begin{aligned}
\bm{\mathcal{A}}_{(i_1, i_2, \cdots, i_d)}
&=\sum_{r_0=r_{d}, r_1,\cdots,r_{d-1}}^{R_0, R_1,\cdots, R_{d-1}}\\
&\bm{\mathcal{G}}^{1}_{(r_{0}, i_1, r_1)}\bm{\mathcal{G}}^{2}_{(r_{1}, i_2, r_2)}\cdots\bm{\mathcal{G}}^{d}_{(r_{d-1}, i_d, r_d)},
\end{aligned}
\label{eq:tr_decomp}
\end{equation}
where $R_0, R_1, \cdots, R_d$ are called ranks and $R_0=R_d$. By treating the latent cores equivalently and capturing complex dependencies, TRD enables compact and flexible representation of high-dimensional data~\cite{xie2024neural}.

\section{Motivation}
\label{sec:motivation}
In this section, we present the motivation for our proposed ultra-light parameter-efficient VLM adaptation on various downstream tasks.
We first investigate the redundancy across adapters, which motivates us to construct an ultra-light adapter based on cross-layer tensor decomposition.
Then, we compare the representation abilities of adapters achieved by different tensor decomposition strategies. 
Finally, we analyze how the rank of the adapter affects the fine-tuning performance based on the chosen tensor decomposition method.

\begin{figure}[t]
    \includegraphics[width=0.48\textwidth]{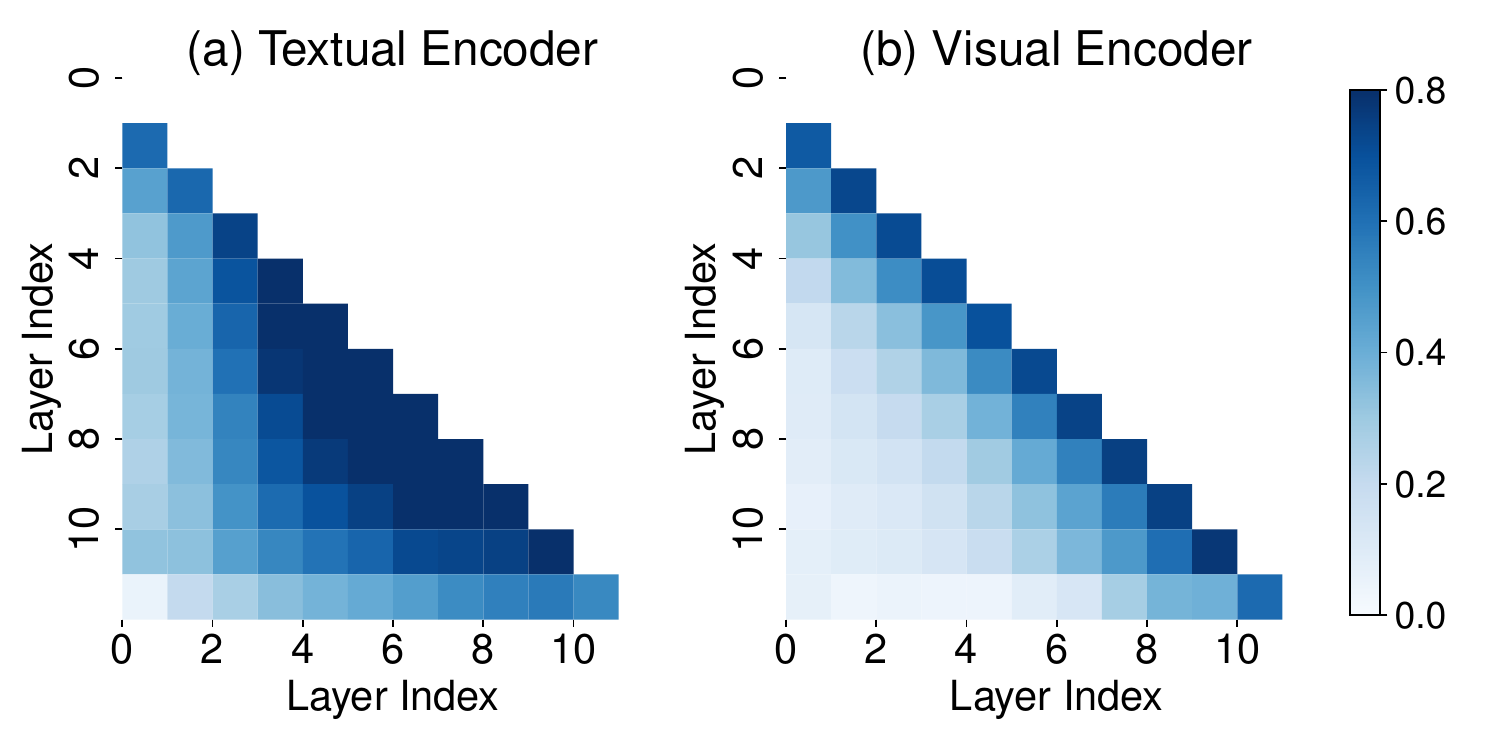}
    \vspace{-5mm}
    \caption{Cosine similarity of embeddings between layers (a) textual encoder (b) visual encoder.}
    \label{fig:obs1_cross_layer}
\end{figure}

\textbf{Observation 1:} \textit{High redundancy exists among adapters across all layers.} In Figure \ref{fig:obs1_cross_layer}, we display the similarities between the embeddings of each layer using data from the OxfordPets dataset~\cite{parkhi2012cats}. The dark color represents high similarities between the embeddings from two different layers. We can observe that cosine similarity exceeds 80\% between two adjacent embeddings. Moreover, a high correlation persists even across layers with large intervals, suggesting that it's parameter-inefficient to employ the layer-level adapters.
To mitigate this redundancy, we propose to exploit the tensor-level low-rankness along the layer dimension by performing tensor decomposition, leading to the following observations.

\begin{figure}[t]
    \includegraphics[width=0.48\textwidth]{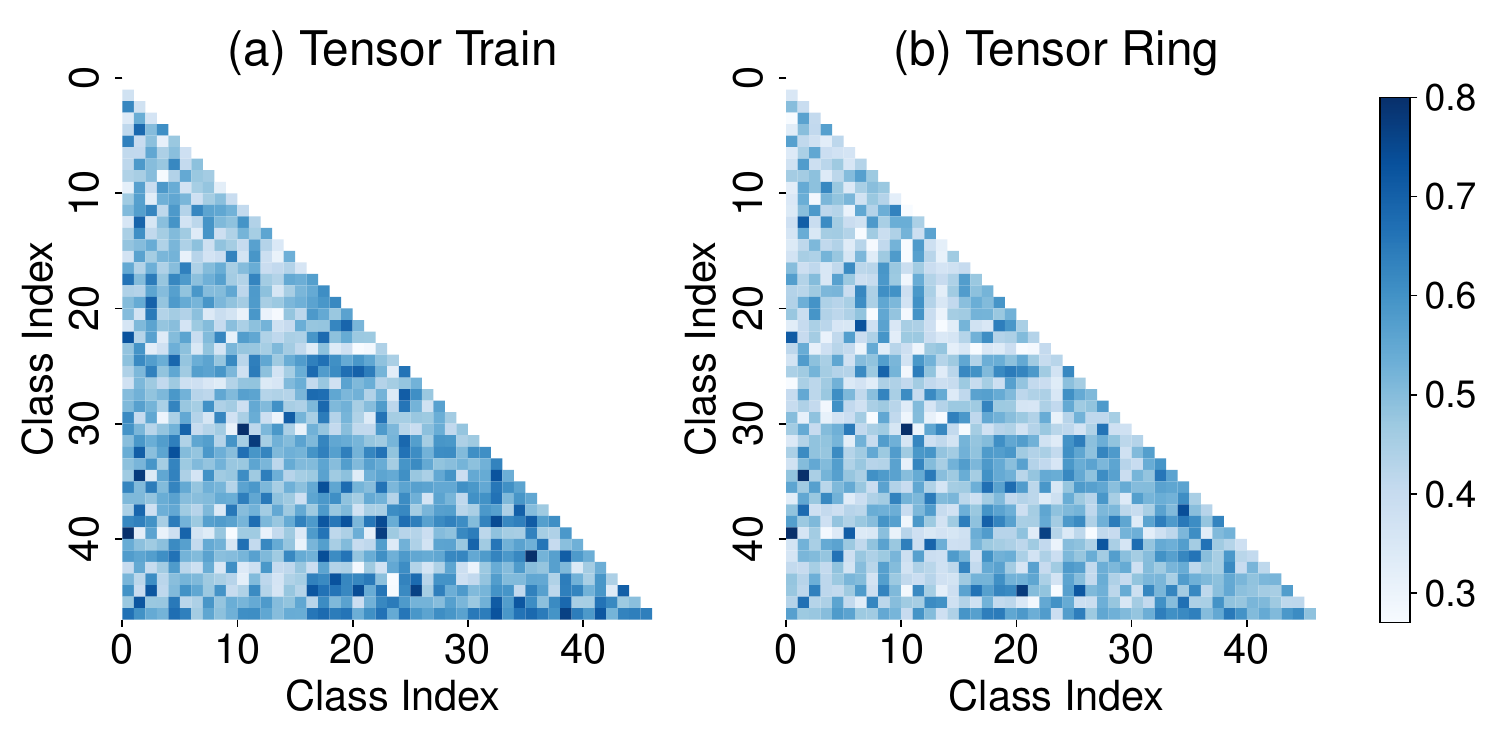}
    \vspace{-5mm}
    \caption{Cosine similarity of textual embeddings with (a) Tensor Train decomposition and (b) Tensor Ring decomposition.}
    \label{fig:tensor_decomp}
\end{figure}

\textbf{Observation 2:} \textit{Adapter with tensor ring structure has more powerful and separate representation abilities, compared to other tensor decomposition methods.}
To select an optimal tensor decomposition strategy that achieves both high compression rate and strong representational abilities, we conduct fine-tuning on the DTD dataset using an adapter achieved by tensor ring decomposition (TRD) ~\cite{zhao2016tensor} and tensor train decomposition (TTD)~\cite{yang2024adazeta}, respectively.
In Figure \ref{fig:tensor_decomp}, we visualize the cosine similarity of textual embeddings for 47 classes. Lighter colors indicate lower similarity. This visualization shows that textual embeddings of different classes produced by the TR adapter exhibit less similarity and greater separation in the feature space, which can reduce the misclassification risk.
We reason that better separability is gained by removing the constraints over ranks and treating the latent cores equivalently within the TRD process, thereby offering more freedom in finding the optimal representations. However, TTD highly depends on permutations of tensor cores, leading to difficulties in finding the optimal representations~\cite{zhao2016tensor}.
Correspondingly, we propose to utilize TRD to achieve an ultra-compact adapter by decomposing the original cross-layer stacked adapters.

\begin{figure}[t]
    \includegraphics[width=0.48\textwidth]{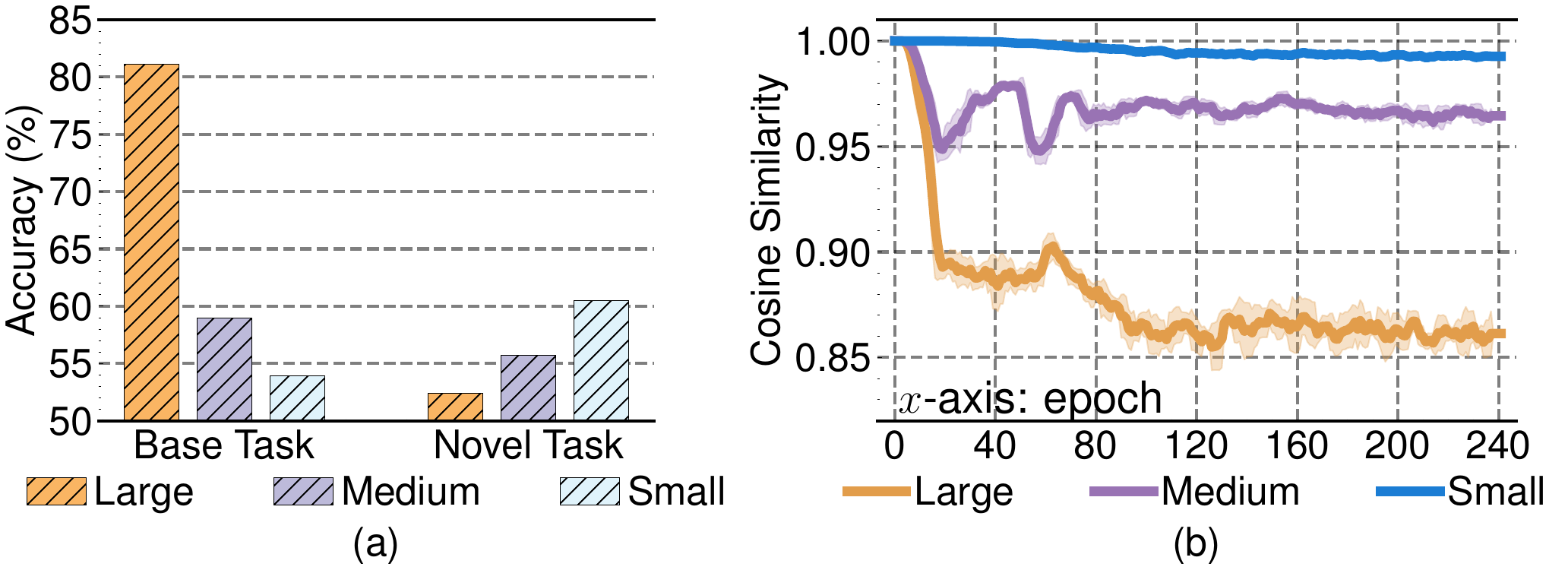}
    \vspace{-5mm}
    \caption{Analysis of diverse adapters. (a) The accuracy in the base and novel tasks. (b) The similarity between the visual embeddings from the frozen CLIP and those from CLIP with fine-tuned adapters.}
    \label{fig:obs3}
\end{figure}

\textbf{Observation 3:} \textit{Adapters with different ranks exhibit different expertise.}
To design adapters that can handle various downstream tasks effectively, we explore two tasks following previous works~\cite{zhou2022conditional, bulat2023lasp}: \uline{1)} base task--testing on seen data; \uline{2)} novel task--testing on unseen data. These tasks emphasize different capabilities: discrimination for the base task and generalization for the novel task. Figure \ref{fig:obs3} (a) shows that CLIP with the large-rank adapter performs better on the base task, reflecting stronger discrimination, while CLIP with the small-rank adapter is more generalizable to the novel task. Additionally, compared to the one with a large-rank adapter in Figure \ref{fig:obs3} (b), CLIP with a small-rank adapter generates visual embeddings that are more similar to those of the frozen CLIP, while the frozen CLIP is proven to have strong generalization abilities that benefited from large-scale pretraining. 
We posit that this divergence is an artifact of parameter count: the large-rank adapter with more parameters is inclined toward capturing discriminative information, while the small-rank adapter with fewer parameters is prone to preserving the generalization ability of a pre-trained model. 
The distinct expertise between adapters offers valuable insights for enhancing both discrimination and generalization, suggesting that adapters of varied ranks can cooperate to handle inputs that require distinct representational capacities.

\section{Methodology: \abbr}

In this section, we introduce the proposed \abbr, an ultra-light parameter-efficient VLM adaptation framework.
We first introduce the cross-layer tensor ring decomposition to address the parameter redundancy among layer-wise adapters in Section~\ref{section:tensor-ring}. Then, we delineate the collaboration of rank-driven adapters for handling various downstream tasks in Section~\ref{section:diverse-adapter}.
Finally, we propose the generalization-aware fine-tuning to further encourage rank-driven adapters to cooperate efficiently in Section~\ref{section:generalization-aware}.

\subsection{Adaptation with Cross-Layer Tensor Ring Decomposition}
\label{section:tensor-ring}
Observation 1 reveals redundancy among layer-wise adapters, while Observation 2 demonstrates that adapters with tensor ring (TR) structure have more powerful and separate representation capabilities.
Therefore, we propose to design cross-layer adapters with the TR structure to reduce fine-tuning parameters, which involves \textit{Cross-Layer Tensorization} and \textit{Ring-based Decomposition}. Finally, the \textit{TR-Structure Adapter} is fine-tuned towards downstream tasks.

\textbf{Cross-Layer Tensorization.} \textit{Tensorization} is used to transform a 2-D weight matrix into a high-dimensional tensor.  
Considering the redundancy among layer-wise adapters, we employ \textit{Cross-Layer Tensorization} for their weight matrices to facilitate low-rankness exploration across layer dimension. Specifically, $L$ adapter weights can be collected and stacked as $\left\{ \bm{A}_i\in \mathbb{R}^{I \times O}\right\}_{i=1}^L$ and then reshaped as $\bm{\mathcal{A}} \in \mathbb{R}^{I_1 \times \cdots \times I_p \times L \times O_1 \times \cdots \times O_q}$, where $I=\prod_{j=1}^p I_j$ and $O=\prod_{j=1}^q O_j$.

\textbf{Ring-based Decomposition.}  After tensorizing these adapters into a high-dimensional tensor $\bm{\mathcal{A}}$, it can be then decomposed into the TR format with $p+q+1$ core tensors $\bm{\mathcal{G}}^{j}$ multiplied one by one, each of which corresponds to an input dimension, layer dimension or an output
dimension, referring to Eq.~\ref{eq:tr_decomp}. The element of tensor $\bm{\mathcal{A}}$ is explicitly calculated as: 
\begin{equation}
\begin{aligned}
& \bm{\mathcal{A}}_{(i_1,  \cdots, i_p, l, o_1, \cdots, o_q)} = \sum_{r_0,\cdots,r_{p+q}} \bm{\mathcal{G}}^{1}_{(r_{0}, i_1, r_1)}\cdots \\
& \bm{\mathcal{G}}^{p}_{(r_{p-1}, i_p, r_p)} \bm{\mathcal{G}}^{p+1}_{(r_p, l , r_{p+1})}\bm{\mathcal{G}}^{p+2}_{(r_{p+1}, o_1, r_{p+2})}\cdots\bm{\mathcal{G}}^{p+q+1}_{(r_{p+q}, o_q, r_{0})}.\\
\end{aligned}
\end{equation}
In particular, $\bm{\mathcal{G}}^{p+1} \in \mathbb{R}^{R_{p} \times L \times R_{p+1}}$ is the decomposed component along layer dimension, with the layer rank parameter $R_{p} (R_{p+1})$  controlling the low-rankness of stacked adapter weight matrices.

\textit{Analysis of training parameters.} 
By exploiting the low-rank property of adapter weights, \abbr~adds only
$R_{p+1}R_{p+2}L + \sum_{j=1}^{p} R_jR_{j+1}I_j + \sum_{j=p+2}^{\,p+q+1} R_jR_{j+1}O_j$ parameters. Since $R_{p+1}$ and $R_{p+2}$ are small values, it alleviates the parameter-efficiency dilemma for the large model with deep layers and wide dimensions. 
In contrast, existing methods ignore the redundancy across adapters, causing the number of training parameters $(I+O)*r*L$ to increase drastically with the increase of layers.

\textbf{TR-Structure Adapter.} 
Based on \textit{Cross-Layer Tensorization} and \textit{Ring-based Decomposition}, AdaRing explicitly determines the structural form of the adapter, referred to as the TR-Structure adapter. It should be noted that the adapter remains in its low-rank tensor format throughout fine-tuning, meaning it is trained to decompose. \abbr~extracts and formulates the redundant components as layer-shared tensor cores. As a result, the layer-wise components can be made substantially smaller, while still retaining the layer-specific characteristics.
Specifically, $\left\{\bm{\mathcal{G}}^{1},\cdots,\bm{\mathcal{G}}^{p},\bm{\mathcal{G}}^{p+2},\cdots,\bm{\mathcal{G}}^{p+q+1}\right\}$ are shared tensor cores among layers to capture the layer-shared information, while the slice $G_l=\bm{\mathcal{G}}^{p+1}_{(:,l,:)}\in \mathbb{R}^{R_p \times R_{p+1}}$  is specific to the adapter $\mathcal{F}_l$ at the $l$-th layer.

We use different initialization strategies for different components of the adapter. Specifically, the layer-specific slice $G_l$ is initialized to zeros to make the initial model’s behavior match that of the original CLIP. Layer-shared tensor cores are initialized from the Gaussian distribution, encouraging the optimizer to explore diverse update directions~\cite{yang2024loretta}.

The forward computation within the adapter $\mathcal{F}_l$ at the $l$-th layer from input embedding $\boldsymbol{x}$ to output embedding $\boldsymbol{y}$ is defined as:
\begin{equation}
\begin{aligned}
\boldsymbol{y} &= \mathcal{F}_l(\boldsymbol{x}). \\
\end{aligned}
\end{equation}
It involves the process of ``\textit{Tensorization-Tensor Multiplication-Vectorization}'':

\textit{\Circled{1} Tensorization.} The input embedding of the $l$-th layer $\boldsymbol{x} \in \mathbb{R}^I$ is first tensorized into the high-dimensional tensor $\bm{\mathcal{X}}\in \mathbb{R}^{I_1 \times \cdots \times I_p}$.

\textit{\Circled{2} Tensor Multiplication.} The input tensor $\bm{\mathcal{X}} \in \mathbb{R}^{I_1 \times \cdots \times I_p}$ then multiplies with layer-shared tensor cores and layer-specific slice sequentially:
\begin{equation}
\begin{aligned}
& \bm{\mathcal{Y}}_{(o_1, \cdots, o_q)} = \sum_{i_1,\cdots,i_p} \sum_{r_0,\cdots,r_{p+q}}
\bm{\mathcal{X}}_{(i_1, \cdots, i_p)} \bm{\mathcal{G}}^{1}_{(r_{0}, i_1, r_1)}\cdots \\
& \bm{\mathcal{G}}^{p}_{(r_{p-1}, i_p, r_p)} G_l\bm{\mathcal{G}}^{p+2}_{(r_{p+1}, o_1, r_{p+2})}\cdots\bm{\mathcal{G}}^{p+q+1}_{(r_{p+q}, o_d, r_{0})}.
\end{aligned}
\end{equation}

\textit{\Circled{3} Vectorization.} Given that the desired output $\boldsymbol{y}$ of the adapter is a vector, the output tensor $\bm{\mathcal{Y}}\in\mathbb{R}^{O_1 \times \cdots \times O_q}$ is finally reshaped into 1-D format $\boldsymbol{y}\in\mathbb{R}^{O}$, where $O=\prod_{j=1}^q O_j$.

\subsection{Collaboration of Rank-Driven Adapters}
\label{section:diverse-adapter}
The selection of layer rank directly affects the capacity of the layer and the number of training parameters. Observation 3 shows that adapters with different ranks have different expertise across tasks. Therefore, \abbr~designs the collaboration of \textit{fine-grained adapter} with relatively large rank and \textit{coarse-grained adapter} with extremely small rank to capture discriminative and generalizable knowledge simultaneously. The overall forward process at the $l$-th layer is expressed as: 

\begin{equation}
\begin{aligned}
\boldsymbol{y}^{v} &= \mathcal{V}_l(\boldsymbol{x}^{v}) + \overline{\alpha} \, \overline{\mathcal{F}}_{l}^{v}(\boldsymbol{x}^{v}) + \hat{\alpha}\hat{\mathcal{F}}_{l}^{v}( \boldsymbol{x}^{v}), \\
\boldsymbol{y}^{t} &= \mathcal{T}_l(\boldsymbol{x}^{t}) + \overline{\beta} \, \overline{\mathcal{F}}_{l}^{t}(\boldsymbol{x}^{t}) + \hat{\beta} \, \hat{\mathcal{F}}_{l}^{t}(\boldsymbol{x}^{t}),
\end{aligned}
\end{equation}
where $\overline{\mathcal{F}}_{l}^{v}$ and $\hat{\mathcal{F}}_{l}^{v}$ are fine-grained and coarse-grained adapters for visual encoder, respectively. The fused weights $\overline{\alpha}$ and $\hat{\alpha}$ are generated by a combinator composed of a single learnable linear layer, enabling the collaboration between adapters in a data-adaptive manner. The adaptive collaboration is also applied to the textual encoder.
\subsection{Generalization-Aware Fine-Tuning}
\label{section:generalization-aware}
To further encourage diverse rank-driven adapters to cooperate towards downstream tasks, we propose the generalization-aware fine-tuning, thereby fully leveraging the different expertise of diverse rank-driven adapters.
Specifically, during fine-tuning, given an image $I$ with the corresponding text prompt ``a photo of a $\{y\}$'', the backbone encoders $\mathcal{V}$ and $\mathcal{T}$ are frozen, while the weights of adapters are learned to maximize cosine similarity between output visual embedding $f^{v}$ and text embedding $f^{t}$ following the cross-entropy loss: 
\begin{equation}
\begin{aligned}
\mathcal{L}_{\text{cls}} = - \sum_{{I} \in \mathbb{I}} \log 
\frac{
    \exp\left( \cos(f^{v}, f^{t}_{y}) / \tau \right)
}{
    \sum_{c=1}^C \exp\left( \cos(f^{v}, f^{t}_{c} / \tau \right)
},
\end{aligned}
\end{equation}
where $C$ is the number of classes and $\tau$ is the temperature. 
 
The cross-entropy loss $\mathcal{L}_{\text{cls}}$ encourages the model to focus on discriminative information from seen data (base task), leading the combinator to favor  the fine-grained adapter. However, since unseen data (novel task) is unavailable during training, the model lacks guidance to generalize well, resulting in the coarse-grained adapter being underutilized. 
To address this, we propose a generalization-aware regularization term that proactively encourages the coarse-grained adapter to participate during training. Specifically, this term minimizes the distance between fine-tuned and frozen CLIP embeddings.
\begin{equation}
\begin{aligned}
\mathcal{L}_{\mathrm{reg}} = \sum_{{I} \in \mathbb{I}} (1-\cos(f^{v}, \mathcal{V}(I))).
\end{aligned}
\end{equation}
The overall training loss is $\mathcal{L} = \mathcal{L}_{\mathrm{cls}} + \lambda \mathcal{L}_{\mathrm{reg}}$, where $\lambda$ is the preservation ratio. By considering these two terms, \abbr~retains strong discrimination while enhancing generalization on tasks.

\begin{table*}[!t]
    \centering
    \vspace{-3mm}
    \caption{Comparison with state-of-the-art methods on the 11 datasets. “Base” and “Novel” are the recognition accuracies on base and novel tasks, respectively. “HM” is the harmonic mean of these two accuracies.}
    \resizebox{0.9\textwidth}{!}{
\begin{tabular}{ll|ccc|ccc|ccc|ccc}

\toprule
& \multirow{2}{*}{Method} &  \multicolumn{3}{c|}{Average}  & \multicolumn{3}{c|}{Caltech101} & \multicolumn{3}{c|}{OxfordPets} & \multicolumn{3}{c}{ImageNet} \\
& & Base & Novel & HM & Base & Novel & HM & Base & Novel & HM & Base & Novel & HM \\
\midrule
\textit{Zero-shot Inference} & CLIP     &   69.34 & 74.22 & 71.70  & 96.84 & 94.00 & 95.40 & 91.17 & 97.26 & 94.12 & 72.43 & 64.18 & 70.22 \\
\midrule
\multirow{6}{*}{\textit{Prompt Tuning}}   & CoOp   &   82.69 & 65.23 & 72.66  & 98.00 & 89.13 & 93.67 & 95.26 & 94.47 & 94.87 & 74.67 & 67.88 & 71.92 \\
&  KgCoOp      & 80.73 & 73.60 & 77.03  & 97.72 & 94.39 & 96.03 & 94.65 & 97.76 & 96.18 & 75.83 & 69.26 & 72.78 \\
 &  MaPLe      & 82.28 & 75.14 & 78.55  & 97.74 & 94.36 & 96.02 & 95.43 & 97.76 & 96.58 & 76.66 & 70.54 & 73.47 \\
 &  LASP       & 82.70 & 74.90 & 78.61  & 98.10 & 94.24 & 96.16 & 95.90 & 97.93 & 96.90 & 76.20 & 70.95 & 73.48 \\
 &  RPO           & 81.13 & 75.00 & 77.78  & 97.97 & 94.37 & 96.03 & 94.63 & 97.50 & 96.05 & 76.60 & 71.57 & 74.00 \\
&  PromptKD & 86.96 & 80.73 & 83.73 & 98.91 & 96.65 & 97.77 & 96.30 & 98.01 & 97.15 & \textbf{80.83} & \textbf{74.66} & \textbf{77.62} \\
\midrule
\multirow{4}{*}{\textit{Adapter Tuning}}   & BSLoRA (SS)  &   82.84 & 74.06 & 78.20  & 98.51 & 94.54 & 96.48 & 94.94 & 97.87 & 96.38 & 77.40 & 70.44 & 73.75 \\
  &     BSLoRA (GT)  &     82.79 & 73.91 & 78.09  & 97.80 & 94.97 & 96.36 & 95.85 & 97.25 & 96.54 & 77.48 & 70.20  & 73.66 \\
 & MMA     &      83.20 & 76.80 & 79.87  & 98.40 & 94.00 & 96.15 & 95.40 & 98.07 & 96.72 & 77.31 & 71.00 & 74.02 \\
\cmidrule(lr){2-14} 
  & \textbf{AdaRing} &  \textbf{87.16}  & \textbf{81.12} &  \textbf{84.03} & \textbf{99.00} & \textbf{96.84} & \textbf{97.91} & \textbf{96.46} & \textbf{98.31} &  \textbf{97.38} & 80.07 & 73.97 & 76.90  \\
\midrule
\midrule
& \multirow{2}{*}{Method} & \multicolumn{3}{c|}{UCF101} & \multicolumn{3}{c|}{EuroSAT}  & \multicolumn{3}{c|}{FGVCAircraft} & \multicolumn{3}{c}{Food101}\\
& & Base & Novel & HM & Base & Novel & HM & Base & Novel & HM & Base & Novel & HM \\
\midrule
\textit{Zero-shot Inference} & CLIP & 70.53 & 77.50 & 73.85 & 56.48 & 64.05 & 60.03& 27.19 & 36.29 & 31.09 &  90.10 & 91.22 & 90.66  \\
\midrule
\multirow{6}{*}{\textit{Prompt Tuning}}   & CoOp  & 84.69 & 56.05 & 67.46 & 92.19 & 54.74 & 68.69 & 40.44 & 22.30 & 28.75 &  88.33 & 82.26 & 85.19  \\
& KgCoOp & 82.89 & 76.67 & 79.65  & 85.64 & 64.34 & 73.48& 36.21 & 33.55 & 34.83 &  90.50 & 91.70 & 91.09  \\
& MaPLe & 83.00 & 78.66 & 80.77  & 94.07 & 73.23 & 82.35 & 37.44 & 35.61 & 36.50 &  90.71 & 92.05 & 91.38  \\
& LASP  & 84.77 & 78.03 & 81.26  & 94.60 & 77.78 & 85.36 & 34.53 & 30.57 & 32.43 &  91.20 & 91.70 & 91.44  \\
& RPO & 83.67 & 75.43 & 79.34  & 86.63 & 68.97 & 76.79 & 37.33 & 34.20 & 35.70 &  90.33 & 90.83 & 90.58  \\
& PromptKD & 89.71 & 82.27 & 86.10 & 97.54 & 82.08 & 89.14& 49.12& 41.81& 45.17 & 92.43 & 93.68 & 93.05\\
\midrule
\multirow{4}{*}{\textit{Adapter Tuning}}   & BSLoRA (SS) & 85.47 &  79.07 & 82.14  & 83.74 & 76.59 & 80.00  & 42.25 & 29.99 & 35.08 & 89.70  & 90.44 & 90.07  \\
  &  BSLoRA (GT) & 85.88 & 76.42 & 80.87  & 83.59 & 77.87 & 80.63 & 39.61 & 27.41 & 32.40 & 89.01 & 90.68 & 89.84 \\
& MMA & 86.23 & 80.03 & 82.20  & 85.46 & 82.34 & 83.87 &40.57 & 36.33 & 38.33 &  90.13 & 91.30 & 90.71  \\
\cmidrule(lr){2-14} 
 &  \textbf{AdaRing} & \textbf{89.93} & \textbf{82.64} & \textbf{86.13}   & \textbf{97.67} & \textbf{85.87} & \textbf{91.39} & \textbf{49.37} & \textbf{41.98} & \textbf{45.38} & \textbf{93.53}  &  \textbf{94.02} & \textbf{93.77}  \\
\midrule
\midrule
& \multirow{2}{*}{Method} & \multicolumn{3}{c|}{SUN397} & \multicolumn{3}{c|}{DTD}  & \multicolumn{3}{c|}{StandfordCars} & \multicolumn{3}{c}{OxfordFlowers} \\
& & Base & Novel & HM & Base & Novel & HM & Base & Novel & HM & Base & Novel & HM \\
\midrule
\textit{Zero-shot Inference} & CLIP & 69.36 & 75.35 & 72.23 & 53.24 & 59.90 & 56.37 & 63.37 & 74.89 & 68.65 &  72.08 & 77.80 & 74.83  \\
\midrule
\multirow{6}{*}{\textit{Prompt Tuning}}   & CoOp & 80.60 & 65.89 & 72.51 & 79.44 & 41.18 & 54.24 &  78.12 & 60.40 & 68.13 & 97.60 & 59.67 & 74.06  \\
& KgCoOp & 80.29 & 76.53 & 78.36 & 77.55 & 54.99 & 64.35 &  71.76 & 75.04 & 73.36 & 95.00 & 74.73 & 83.65  \\
& MaPLe & 80.82 & 78.70 & 79.75 & 80.36 & 59.18 & 68.16 & 72.94 & 74.00 & 73.47  &  95.92 & 72.46 & 82.56 \\
& LASP & 80.70 & 78.60 & 79.63 & 81.40 & 58.60 & 68.14 &  75.17 & 71.60 & 73.34 & 97.00 & 73.53 & 83.95  \\
& RPO & 80.60 & 77.80 & 79.18 & 76.70 & 62.13 & 68.61 &  73.87 & 75.53 & 74.69 & 94.13 & 76.67 & 84.50  \\
& PromptKD & 83.69 & 81.54 & 82.60 & 85.84 & 71.37 & 77.94 & 82.80 & 83.37 & 83.13 & 99.42 & \textbf{82.62} & \textbf{90.24} \\
\midrule
\multirow{4}{*}{\textit{Adapter Tuning}}   & BSLoRA (SS) & 81.75 & 75.88 & 78.70  & 82.52 & 58.45 & 68.43 & 77.64 & 74.58 & 76.08 & 97.34 & 66.80 & 79.23 \\
  &  BSLoRA (GT) & 81.84 & 75.84  & 78.73  & 83.21 & 56.15 & 67.05  & 79.03 & 74.44 & 76.66 & 97.34 & 71.77 & 82.62 \\
  & MMA & 82.27 & 78.57 & 80.38 & 83.20 & 65.63 & 73.38  & 78.50 & 73.10 & 75.70 & 97.77 & 75.93 & 85.48 \\
\cmidrule(lr){2-14} 
 & \textbf{AdaRing} & \textbf{83.72} & \textbf{81.99} &  \textbf{82.85} & \textbf{86.04} & \textbf{72.39} & \textbf{78.63}&  \textbf{83.47} & \textbf{84.05} & \textbf{83.76}  &\textbf{99.51} & 80.25 & 88.85 \\
\bottomrule
\end{tabular}}
 
    \label{tab:overall_result}
\end{table*}

\section{Experiments}

\textbf{Datasets.} Following~\cite{yang2024mma}, we conduct experiments on 11 image classification datasets, i.e., ImageNet~\cite{deng2009imagenet} and Caltech~\cite{fei2004learning} for generic object classification; OxfordPets~\cite{parkhi2012cats}, StanfordCars~\cite{krause20133d}, OxfordFlowers~\cite{nilsback2008automated}, Food101~\cite{bossard2014food}, and FGVCAircraft~\cite{maji2013fine} for fine-grained visual categorization, EuroSAT~\cite{helber2019eurosat} for satellite image classification, UCF101~\cite{soomro2012ucf101} for action recognization, DTD~\cite{cimpoi2014describing} for texture classification, and SUN397~\cite{xiao2010sun} for scene understanding.

\textbf{Baselines.} On all datasets, we compare our proposed \abbr~with zero-shot CLIP~\cite{radford2021learning}, prompt tuning approaches including CoOp~\cite{zhou2022learning}, KgCoOp~\cite{yao2023visual}, MaPLe~\cite{khattak2023maple}, LASP~\cite{bulat2023lasp}, RPO~\cite{lee2023read}, PromptKD~\cite{li2024promptkd} and existing state-of-the-art adapter-based fine-tuning approaches, BSLoRA~\cite{zhou2025bslora}, MMA~\cite{yang2024mma}.

\textbf{Implementation Details.}
Following previous works, we conduct experiments with the few-shot setting, i.e. 16 shots per category. We use ViT-B/16 based CLIP model in all settings of experiments and train our models for 10 epochs. On the large-scale ImageNet dataset, we use a batch
size of 128 for training. On the other 10 datasets, we set the batch size to 16.

\subsection{Main Results}
We compare \abbr~ with state-of-the-art methods on the 11 widely used datasets of base and novel tasks, as well as harmonic mean (HM) of accuracies on base and novel tasks. As illustrated in Table \ref{tab:overall_result}, our \abbr~ demonstrates superior average performance across 11 datasets, ranking first on base and novel tasks in 9 of them. Notably, on the EuroSAT dataset, \abbr~attains 95.04\% on the base task and 85.87\% on the novel task, outperforming MMA, the adapter tuning-based approach with the second-best average performance, by 12.21\% and 3.53\%, respectively.
Moreover, Bi-Share (GT) exhibits a significant performance drop on the fine-grained dataset FGVCAircraft. We argue that it's because the lossy process of one-rank decomposition to gates hinders the model from capturing fine-grained information.
These results show that our approach achieves both discriminative and generalizable performance in varied downstream tasks.

\textbf{Efficiency Comparison.} In addition to performance evaluation, we also compare the training parameters of our method \abbr, with existing VLM fine-tuning methods in the OxfordFlowers dataset.
In Table \ref{tab:efficiency_comparison}, \uline{compared to MMA, \abbr~ achieves a significant reduction in terms of training parameters by approximately 90\%.} Moreover, PromptKD relies on a computationally expensive two-stage process by first finetuning a large CLIP teacher model and then distilling the knowledge from the teacher model to the small student model. Besides, our AdaRing maintains a comparable parameter scale to that of CoOp and LASP while achieving significantly superior performance. These improved performances are induced by the components of the cross-layer tensor ring decomposition and diverse adapters in \abbr, which will be proved in the ablation studies.

\begin{table*}[!h]
    \centering
    
    \caption{Efficiency comparison in terms of learnable parameters.}
    \vspace{-3mm}
    \resizebox{0.8\textwidth}{!}{%

\begin{tabular}{ccccccccc}
  \toprule[1.5pt]
Method & CoOp & MaPLe & LASP & PromptKD & BSLoRA (SS) & BSLoRA (GT) & MMA & \textbf{AdaRing} \\
  \midrule[0.75pt]
 \#Param. (M) & 0.04 & 3.55 & 0.06 & 1.01 & 0.15 & 0.21 & 0.67 & \textbf{0.06} \\
  \bottomrule[1.5pt]
\end{tabular}%
} 
    \label{tab:efficiency_comparison}
\end{table*}

\begin{table*}[!h]
    \centering
    \caption{Ablation study on the main components of \abbr.}
    \vspace{-3mm}
    \resizebox{0.8\textwidth}{!}{
\begin{tabular}{c|ccc|cc|cc|cc}
\toprule[1.5pt]
\multicolumn{1}{l|}{\multirow{2}{*}{Method}} & \multirow{2}{*}{\makecell{Cross\\Layer}} & \multirow{2}{*}{\makecell{Diverse\\Adapter}} & \multirow{2}{*}{\makecell{\# Param. \\(M)}} & \multicolumn{2}{c|}{FGVCAircraft}     & \multicolumn{2}{c|}{OxfordPets}& \multicolumn{2}{c}{Food101}                          \\
\multicolumn{1}{l|}{}                         &                              &                                  &                                                        &  Base       & Novel     &  Base       & Novel      & Base & Novel \\ 
\midrule[0.75pt]
\multirow{3}{*}{AdaRing} & \xmark  & \xmark  &0.67  & 40.57 & 36.33 &  95.40 & 98.07  & 90.13 & 91.30   \\
  &  \cmark  &  \xmark &    0.29   & 42.83 &   36.77  &  95.45  & 98.00    & 90.24 & 91.10   \\
  &  \cmark & \cmark & 0.06 & \textbf{49.37} & \textbf{41.98} &  \textbf{96.46} & \textbf{98.31} &  \textbf{93.53}& \textbf{94.02}    \\ 
\bottomrule[1.5pt]
\end{tabular}}
 
    \label{tab:ablation_main_components}
\end{table*}

\subsection{Ablation Studies}
\label{sec:ablation}
\textbf{Ablation of Main Components.} To investigate the importance of cross-layer tensor ring decomposition (TRD) and the collaboration of diverse adapters, we conduct ablation studies on 3 datasets: FGVCAircraft, OxfordPets, and Food101 in Table \ref{tab:ablation_main_components}, 
and present the results of MMA in row 1 as baseline. By comparing row 2 to row 1, \abbr~achieves a substantial reduction in parameter count by nearly 68\% without sacrificing performance, demonstrating the effectiveness of our cross‐layer TRD design.
Additionally, built upon the cross-layer TRD design,  our diverse adapter design achieves better performance with fewer training parameters compared to its homogeneous counterpart (row 3 vs. row 2), demonstrating its efficiency in handling diverse tasks.

\begin{figure}[!t]
    \centering
    \begin{subfigure}{0.48\linewidth}
        \centering
        \includegraphics[width=\linewidth]{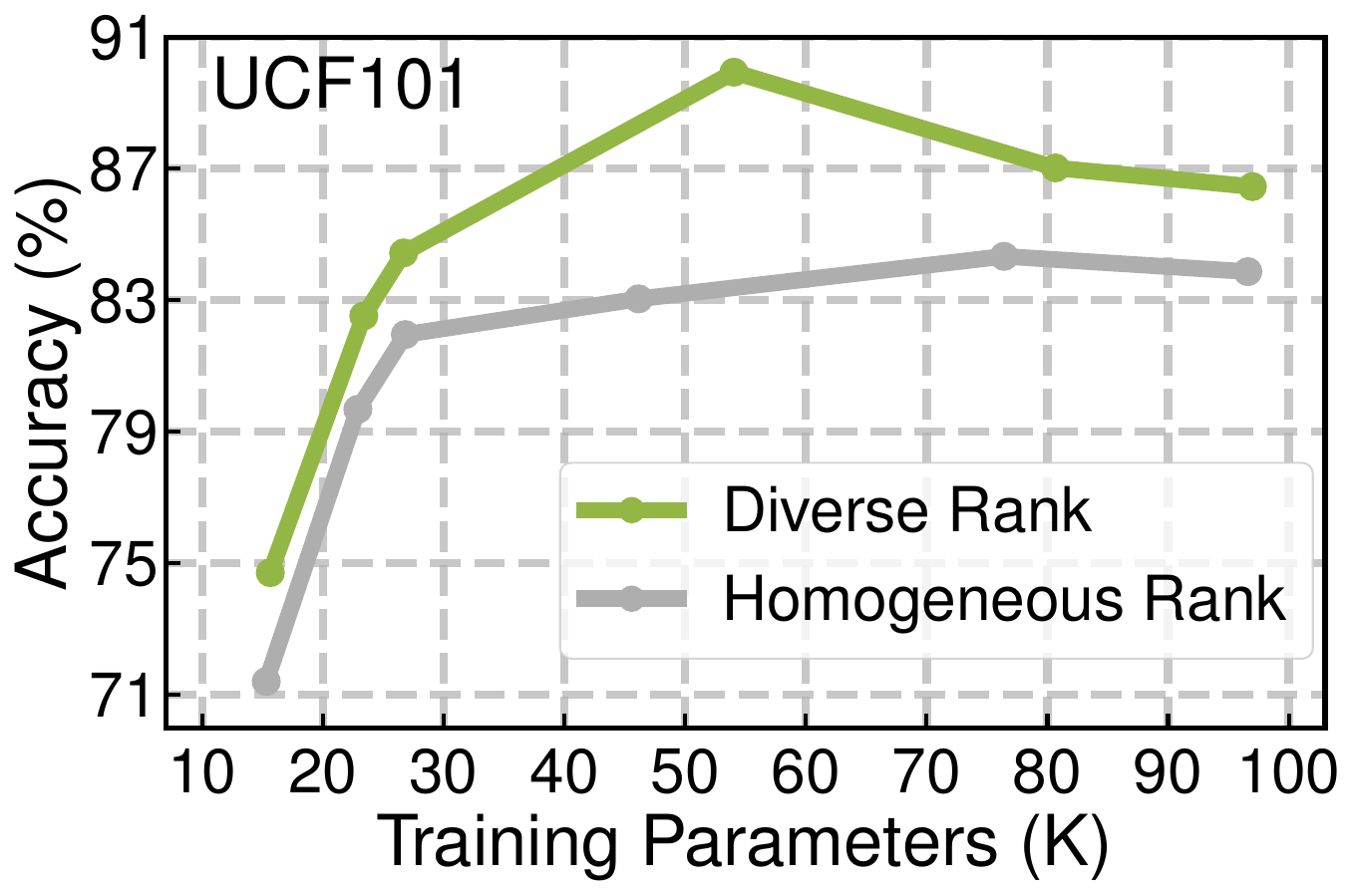}
        \caption{Base Task}
        \label{fig:rank_analysis_subfig1}
    \end{subfigure}
    \hfill
    \begin{subfigure}{0.48\linewidth}
        \centering
        \includegraphics[width=\linewidth]{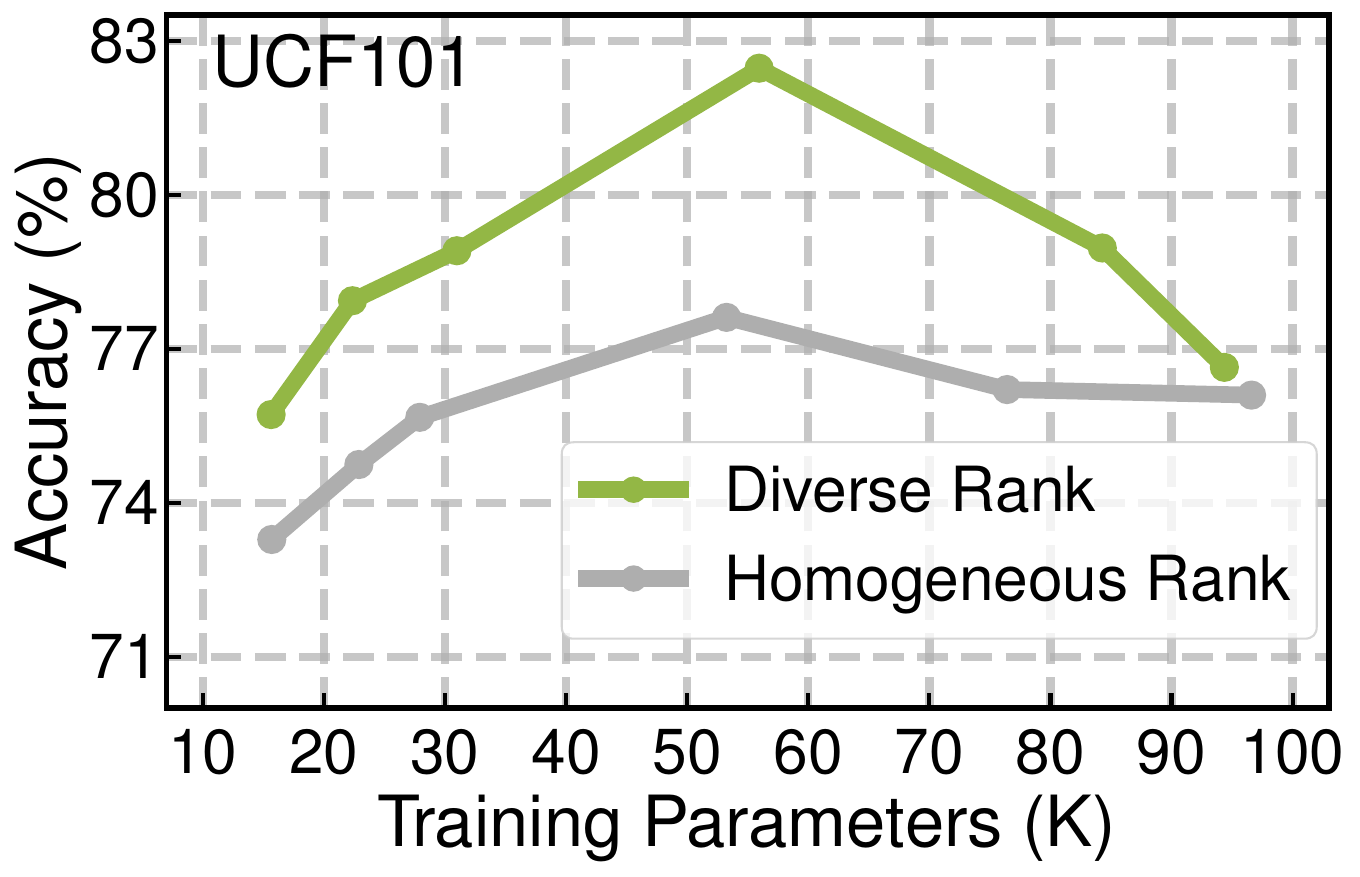}
        \caption{Novel Task}
        \label{fig:rank_analysis_subfig2}
    \end{subfigure}

    \vskip\baselineskip 

    \begin{subfigure}{0.48\linewidth}
        \centering
        \includegraphics[width=\linewidth]{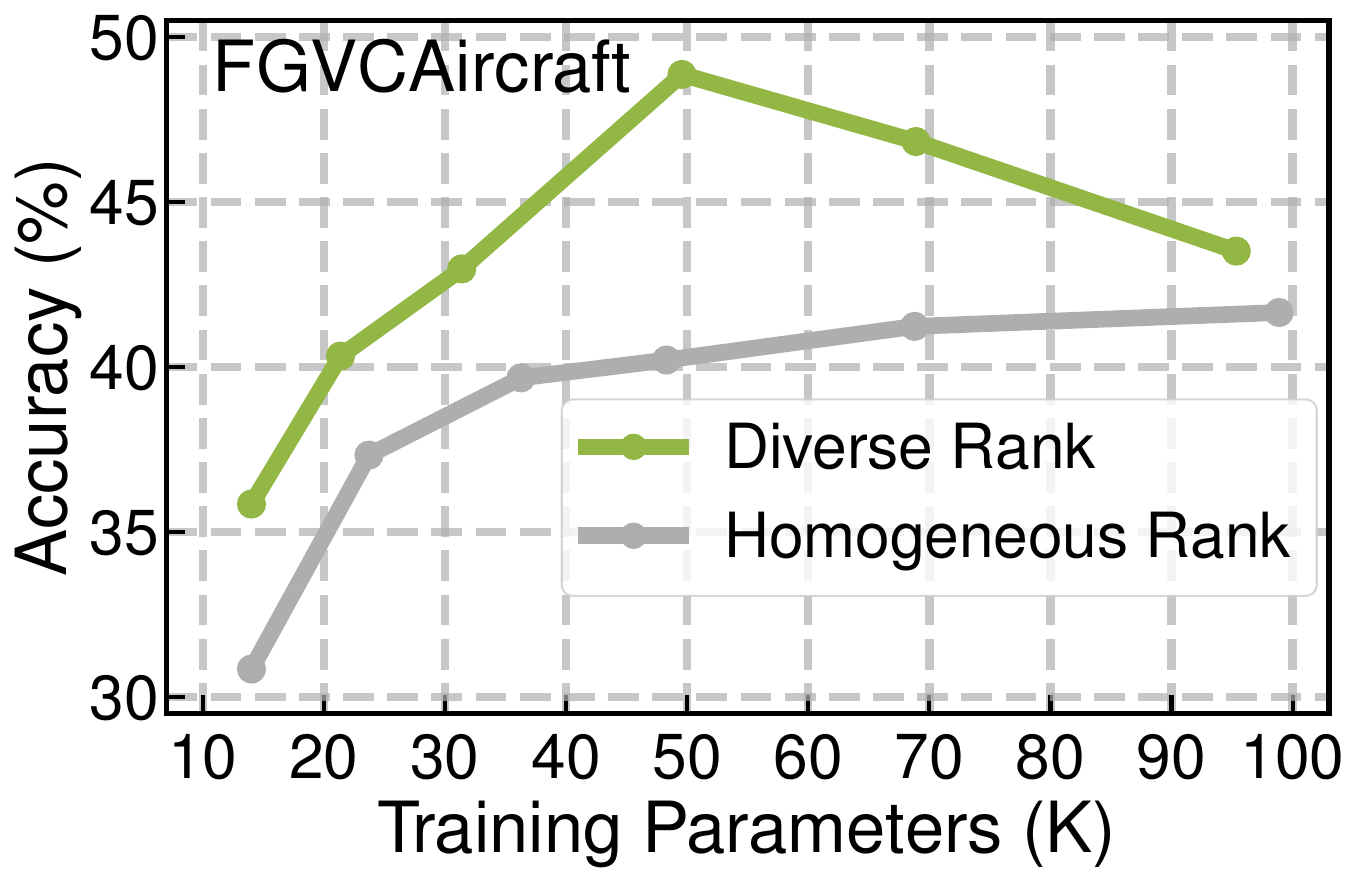}
        \caption{Base Task}
        \label{fig:rank_analysis_subfig3}
    \end{subfigure}
    \hfill
    \begin{subfigure}{0.48\linewidth}
        \centering
        \includegraphics[width=\linewidth]{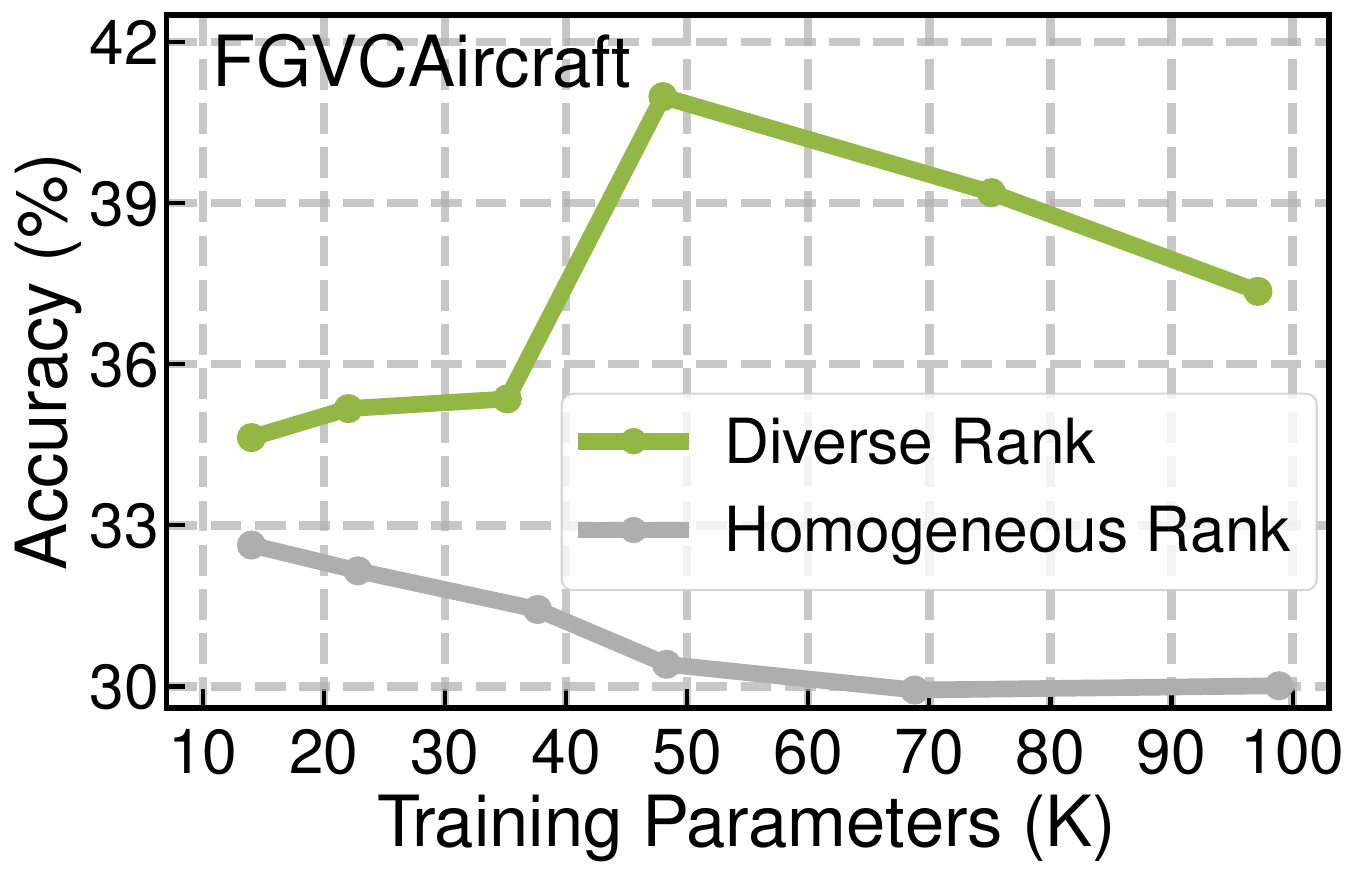}
        \caption{Novel Task}
        \label{fig:rank_analysis_subfig4}
    \end{subfigure}
    \caption{Impact of adapter diversity on performance under varying trainable parameter budgets.}
    \label{tab:rank_analysis}
\end{figure}

\textbf{Diversity of Adapters.} 
To further rigorously examine the impact of diversity in the adapter structure, we compare our diverse adapter design with the homogeneous adapter strategy within a certain range of parameters, by evaluating their performance on the base and novel tasks of UCF101 and FGVCAircraft datasets. 
As Figure \ref{tab:rank_analysis} shown, 
with the increase of training parameters, the accuracy rises to a peak and then decreases for the base and novel tasks.
More importantly, the model with diverse adapters consistently outperforms one with homogeneous adapters under the same number of training parameters.
Specifically, we set the layer rank to 64 in the fine-grained adapter and the layer rank to 1 in the coarse-grained adapter.

\begin{figure}[h]
    \centering
    \includegraphics[width=0.35\textwidth]{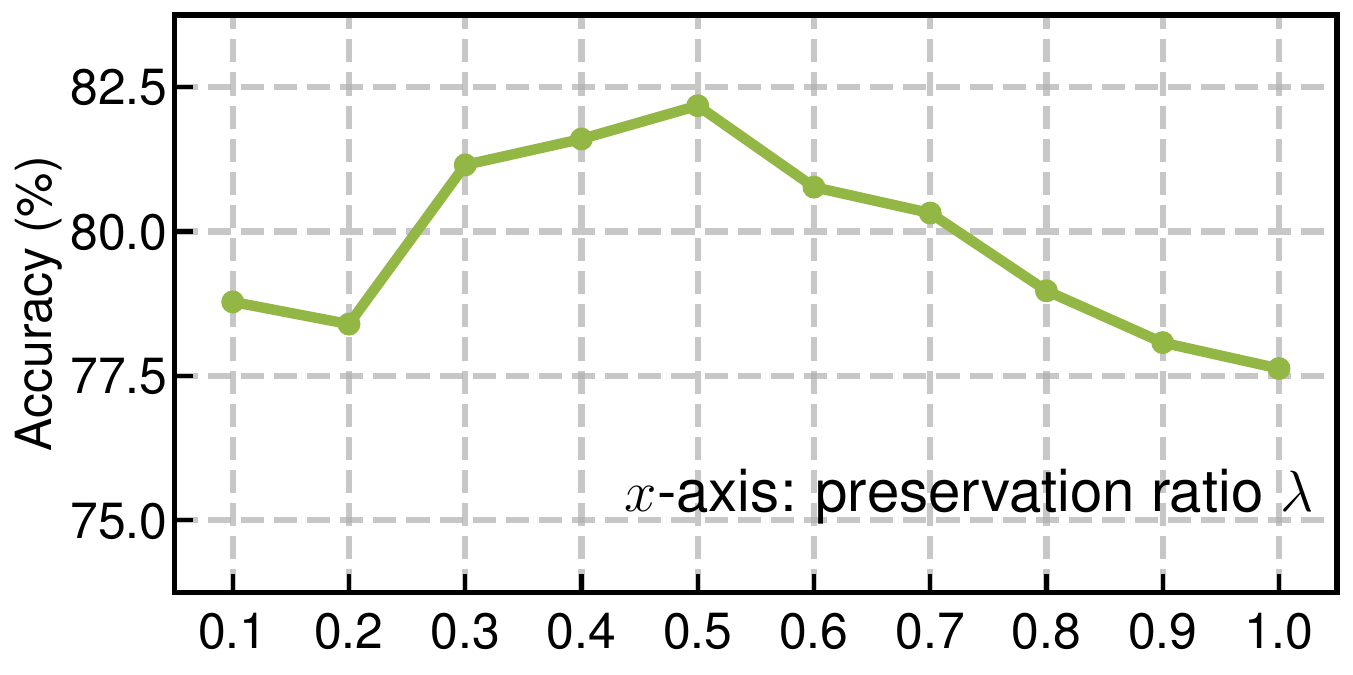}
    \caption{Influence of different preservation ratios on performance of novel task.}
    \label{fig:parameter_analysis}
\end{figure}

\textbf{Hyperparameters Analysis.} Preservation ratio $\lambda$ is a parameter controlling the balance of classification loss and generalization-aware regularization term during fine-tuning. To determine the appropriate setting, we explore the influence of $\lambda$ on model performance in the EuroSAT.
Figure \ref{fig:parameter_analysis} shows that the accuracy on the novel task rises with the increase of $\lambda$. This rise can be attributed to the augmented generalization capacity of the model. 
However, we can observe that performances begin to decrease when the preservation ratio is larger than 0.5. This is because increasing the weight of the generalization-aware regularization term reduces the emphasis on the classification loss during training. Therefore, to achieve an optimal balance between classification accuracy and generalization regularization, we set $\lambda = 0.5$.

\section{Conclusion}

In this work, we propose a novel fine-tuning framework for extremely parameter-efficient vision \& language model adaptation.  Our approach \abbr~constructs the ultra-light adapter into layer-shared and layer-specific components based on the cross-layer tensor ring decomposition algorithm, addressing the high redundancy among layer-wise adapters. Moreover, the collaboration of adapters with varied granularity is developed to handle tasks that require distinct representational capacities. Finally, the diverse adapters are further encouraged to cooperate during the generalization-aware fine-tuning.
Experiments on various downstream tasks demonstrate the superiority of our proposed \abbr.

\clearpage
{
    \small
    \bibliographystyle{ieeenat_fullname}
    \bibliography{main}
}


\end{document}